\pdfoutput=1

\documentclass[11pt]{article}

\usepackage[]{acl}

\usepackage{times}
\usepackage{latexsym}

\usepackage[T1]{fontenc}

\usepackage[utf8]{inputenc}

\usepackage{microtype}

\usepackage[textsize=scriptsize]{todonotes}

\usepackage{enumitem}
\usepackage{booktabs}
\usepackage{multirow}
\usepackage{amsmath}
\usepackage{amsfonts}
\usepackage{amssymb}
\usepackage{fontawesome}
\usepackage[ruled,vlined]{algorithm2e}
\usepackage{arydshln}
\usepackage[procnames]{listings}
\usepackage{multicol}
\usepackage[title]{appendix}
\usepackage{graphicx} 
\usepackage{subcaption}
\usepackage{float}
\usepackage{colortbl}

\definecolor{mygreen}{HTML}{00786C}
\definecolor{lightgrey}{HTML}{447777}
\definecolor{mypurple}{HTML}{D64C1D}

\newcommand{\EG}[1]{\textcolor{mygreen}{#1}}
\newcommand{\CXT}[1]{\textcolor{mypurple}{#1}}

\definecolor{mygray}{gray}{0.4}

\title{Entity Cloze By Date: 
What LMs Know About Unseen Entities}

\author{
Yasumasa Onoe,
Michael J.Q. Zhang,
Eunsol Choi,
Greg Durrett\\
Department of Computer Science\\
The University of Texas at Austin \\
{\tt\{yasumasa, mjqzhang, eunsol, gdurrett\}@cs.utexas.edu}}

\begin{document}
\maketitle
\begin{abstract}
Language models (LMs) are typically trained once on a large-scale corpus and used for years without being updated. However, in a dynamic world, new entities constantly arise. We propose a framework to analyze what LMs can infer about new entities that did not exist when the LMs were pretrained. We derive a dataset of entities indexed by their origination date and paired with their English Wikipedia articles, from which we can find sentences about each entity. We evaluate LMs' perplexity on masked spans within these sentences. We show that models more informed about the entities, such as those with access to a textual definition of them, achieve lower perplexity on this benchmark. Our experimental results demonstrate that making inferences about new entities remains difficult for LMs. Given its wide coverage on entity knowledge and temporal indexing, our dataset can be used to evaluate LMs and techniques designed to modify or extend their knowledge. Our automatic data collection pipeline can be easily used to continually update our benchmark. 

  
\end{abstract}

\section{Introduction} 
New entities arise every day: new movies, TV shows, and products are created, new events occur, and new people come into the spotlight. Whatever the capabilities of language models (LMs) to represent entity knowledge, these new entities cannot possibly be included in the language models' parametric knowledge (i.e., knowledge acquired during pretraining), as they did not exist when LMs were trained. Since this temporal mismatch between LMs and real-world knowledge affects model performance on downstream tasks \citep{zhang-choi-2021-situatedqa, Bhuwan_Dhingra_21, Angeliki_Lazaridou_21}, understanding what LMs know about real-world entities is an important task.


\begin{figure}
    \centering
    \includegraphics[width=1.0\linewidth]{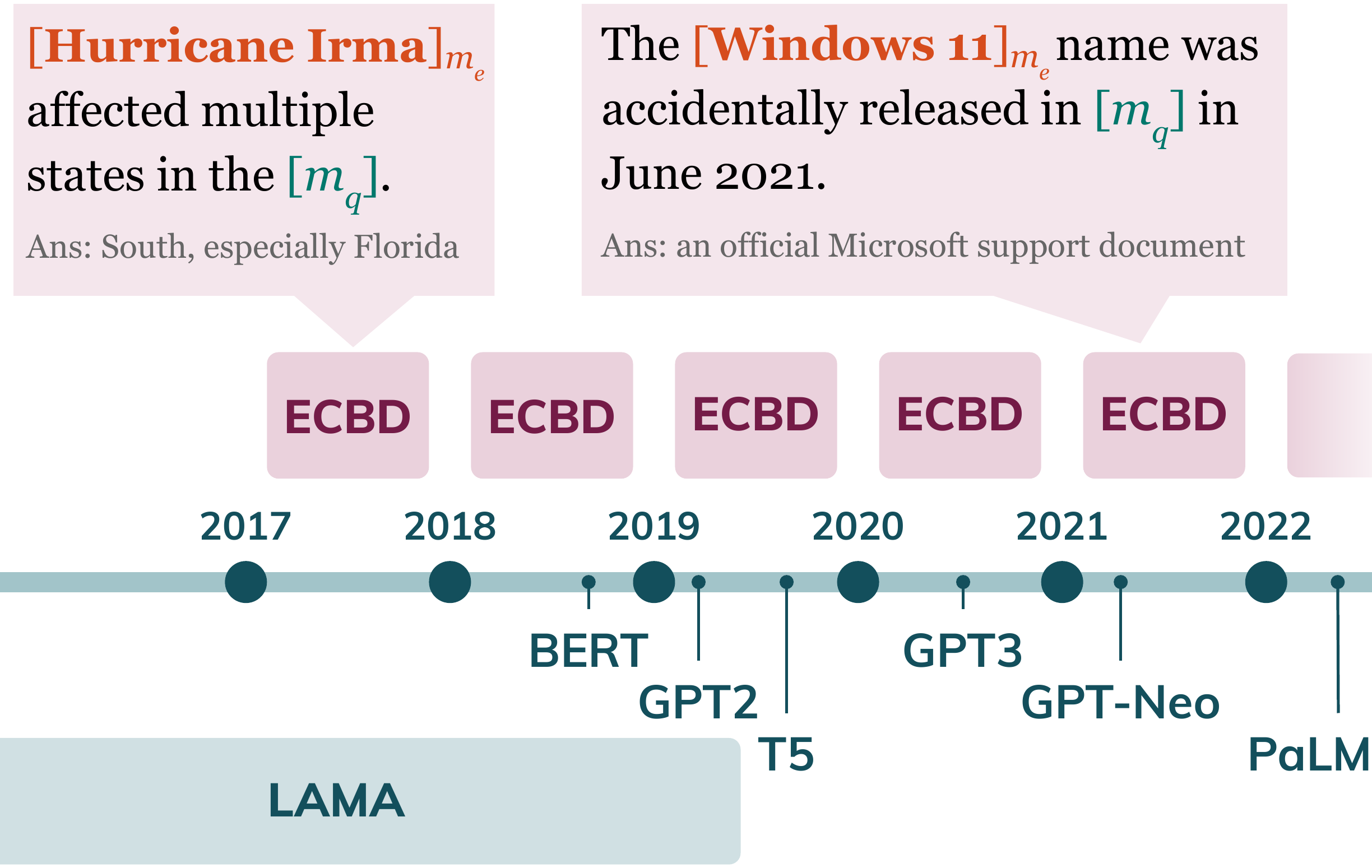}
    \caption{Our framework (ECBD) collects entities indexed by the year when they were first introduced in Wikipedia and their cloze sentences, unlike existing cloze datasets (LAMA~\cite{petroni-etal-2019-language}) which broadly cover entities introduced prior to 2019.}
    \label{fig:data_collection}
    \vspace{-12pt}
\end{figure}


The existing literature provides various benchmarks to measure LMs' knowledge about entities \citep{petroni-etal-2019-language,petroni-etal-2021-kilt,Bhuwan_Dhingra_21}. Those benchmarks are typically formulated as cloze-style tasks covering a limited set of relations bounded by knowledge bases: LAMA uses around 40 Wikidata relations and entities collected in 2017. Newer cloze benchmarks~\citep{Bhuwan_Dhingra_21,Joel_Jang_21} integrate temporal aspects to identify a time period when a cloze sentence is valid, but do not differentiate new and existing entities. These knowledge probing datasets fail to test broad knowledge about real-world entities or evaluate how LMs' knowledge differs on entities that are seen or unseen during pre-training.

To fill this gap, we propose a framework to evaluate LMs' knowledge about entities classified by their origination date. We extract a set of Origination Date Indexed Entities (ODIE) based on metadata from Wikidata. We then construct cloze statements by masking sentences in those entities' Wikipedia articles. Unlike past knowledge probing datasets, these cloze sentences test the ability of a model to make a wide range of inferences related to entities, without being resticted to a pre-defined set of KB relations. We choose masked spans near these entities that likely contain information related to the entities, which we evaluate based on the perplexity gap between the raw sentence and the sentence with the entity replaced. 

We release the Entity Cloze by Date (ECBD) dataset of 35k masked sentences that contain mentions of 2.1K ODIE entities,\footnote{ The code and data are publicly available at \url{https://github.com/yasumasaonoe/ecbd}.} split by year covering a time period from 2017 to 2021, together with 8k masked sentences of popular entities from any time period. In our experiments, we evaluate three pre-trained language models in terms of perplexity. 
We establish that injecting additional information such as a text definition can meaningfully teach the model to make better guesses about masked spans, highlighting this dataset's utility for benchmarking methods of knowledge injection.


\section{Entity Cloze by Date}
We aim to test language models' 1) broader entity knowledge and 2) ability to reason about completely unseen entities (i.e., unseen during pretraining). Thus, we want to have the following properties in our entity cloze sentences. \textbf{(1) Date indexing.} If each cloze example is associated with an entity and indexed by the origination date of that entity, we can understand whether a model may have seen it in its pre-training corpus or not. \textbf{(2) Diverse sentences.} When going beyond KB triples, entity knowledge can take many forms: actions that an entity can take, other entities that action can effect, typical ways in which an entity is described, and more. Thus, we want include diverse sentences and masked spans that cover rich relations and various syntactic categories (e.g., POS and nonterminal categories, span length).


\subsection{Task Definition}
Each entity \CXT{$e$} is paired with \CXT{$e_i$}, its origination year. Given a sentence $s$ containing an entity mention span \CXT{$m_e$} and a masked query span \EG{$m_q$}, a language model is asked to predict the gold masked span \EG{$m_y$}. See the following example: 

\begin{quote}
\CXT{$e$: RNA vaccine, $e_i$: 2020}\\
$s$: \CXT{[mRNA vaccines]$_{m_e}$} do not affect or reprogram \EG{[$m_q$]}. \\
$m_y$: \EG{DNA inside the cell}
\end{quote}
We evaluate language models by \textbf{perplexity} on the masked span \EG{$m_q$} (see Appendix~\ref{sec:appendix_recall} for a discussion of recall as another metric).




\begin{figure}[!t]
    \centering
    \includegraphics[width=1.0\linewidth]{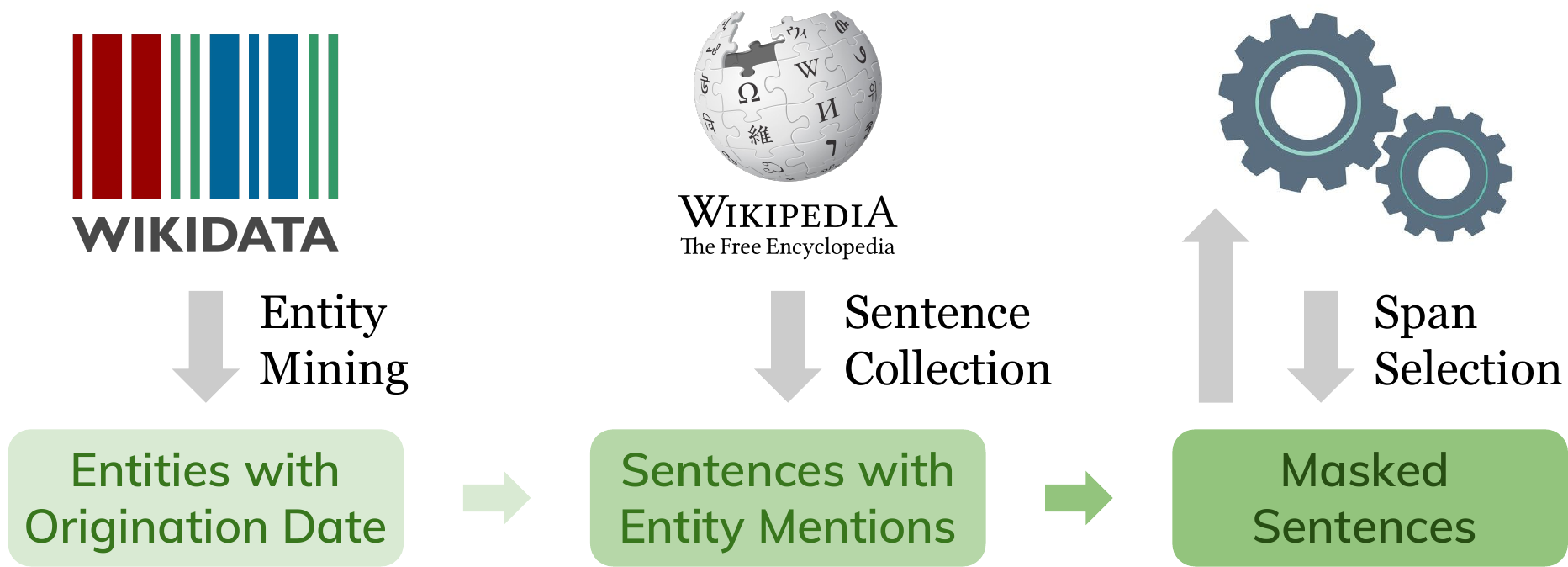}
    \caption{Overview of the data collection process.}
    \label{fig:}
    \vspace{-11pt}
\end{figure}


\subsection{Data Collection}

Our data collection protocol consists of three stages: \emph{entity mining}, \emph{sentence collection} and \emph{span selection}. We use English Wikipedia (the September 1, 2021 dump) and Wikidata as knowledge sources. 
\begin{table*}
	\centering
	\footnotesize
	\setlength{\tabcolsep}{4pt}
	\begin{tabular}{l r r r r r r  l}
		\toprule
		Origination Year              & 2017 & 2018 & 2019 & 2020 & 2021 & Total & \multirow{3}{*}{Example Entities}\\
		\cmidrule(r){1-7}
		\# Dev Entities   &  300 &  280 &  219 &  187 &   78 & 1,050 & \\
		\# Test Entities  &  299 &  279 &  208 &  176 &   80 & 1,029 & \\ \midrule
		Sports            &   20 &   19 &   22 &   12 &   27 &    19 & 2017 Tour de France, USL League One, Evo 2017 \\
		Media             &   18 &   19 &   24 &   23 &   20 &    21 & Emily in Paris, Luigi's Mansion 3, The Midnight Gospel \\
		Infrastructure    &   10 &    8 &   10 &    8 &    9 &     9 & Gateway Arch National Park, Istanbul Airport, I-74 Bridge \\
		Natural Risks     &    3 &    6 &    4 &   15 &   11 &     7 & Hurricane Ida, COVID-19, North Complex Fire\\
		Products          &    4 &    4 &    4 &    3 &    3 &     4 & Apple Card, Sputnik V COVID-19 vaccine, Pixel 4 \\
		Businesses        &   15 &   11 &    7 &    7 &    3 &    10 & Raytheon Technologies, Electrify America, Good Party \\
		Organizations     &   16 &   18 &   13 &   12 &    9 &    15 & NUMTOT, UK Student Climate Network \\
		Other Events      &    9 &   10 &   11 &   12 &   13 &    11 & Super Bowl LIV halftime show, Storm Area 51\\
 		Misc.             &    5 &    3 &    4 &    7 &    4 &     4 & RNA vaccine, Earthshot Prize, Comet NEOWISE \\
		\bottomrule 
	\end{tabular}\vspace{-4pt}
	\caption{Origination date indexed entity (ODIE) statistics by category. The number represents \% of entities with particular type among entities originated in that year. }
	\label{tab:per-type}
	\vspace{-11pt}
\end{table*}

\vspace{-4pt}
\paragraph{ODIE Mining}
We begin by gathering all entities on Wikidata that have an associated \textit{start time}, \textit{announcement date}, \textit{time of discovery or invention}, \textit{inception} date, \textit{point in time}, or date it was \textit{introduced on}. For such entities, we take the first of these dates to create our temporal splits, assuming that this is the earliest date the entity could have appeared in any pretraining corpus. 

To compare with ODIE which covers relatively new entities originated in 2017 at the earliest, we use a set of \textsc{Popular} entities ranked by article contributor numbers and incoming links from prior work~\cite{onoe2021creak,Geva2021DidAU}. 

\vspace{-4pt}
\paragraph{Entity Sentence Collection} Once we obtain a list of entities, we look up their English Wikipedia articles. To enrich the candidate sentence pool and exclude trivial sentences from stub articles, we filter entities if their corresponding articles contain less than 500 words. From each article, we exclude the first paragraph of the article, to be used as an entity definition, and sample sentences from the rest of the paragraphs. We sample sentences that include the entity name or one of their Wikidata aliases. We do not accept entity mention spans located in quotes since they are often in nested named entities such as book titles. 
We also filter out any sentences with less than five words. 

\vspace{-4pt}
\paragraph{Span Selection} Next, we determine spans $m_q$ to be masked on a sentence, $s$; we can have multiple masked spans per sentence, masked separately. All spans must be: (a) not overlapping with the entity mention span, $m_e$, (b) located after the entity mention span, $m_e$, and (c) starting no more than ten words away from the mention span, to improve relatedness to the entity. We select spans after the entity mention so left-to-right language models will condition on the entity at test time.

We extract two types of spans: \textbf{NP spans} are selected from any suitable noun phrases in the sentence using spaCy \citep{spacy2}. These spans primarily represent relational knowledge about the entity, analogous to the object in a KB triple. \textbf{Random spans} are arbitrary sequences of words sampled from the sentence. This broader set of spans may cover other types of entity knowledge (e.g., probable actions an entity can take). We uniformly sample span length between 1 and 5 and then randomly select the starting location of the span within the sentence. We only accept valid spans not overlapping with the entity mention. We extract at most 100 spans per entity to limit any one entity's contribution to the final dataset.

\vspace{-4pt}
\paragraph{Span sensitivity to entity knowledge}
To see if our design choices are effective, we perform a test that measures the performance drop in perplexity using T5 when we replace the entity mention with a generic reference to ``the entity.'' We use entities from our \textsc{Popular} set to ensure that the LM has seen them during pre-training. If a masked span is related to the entity, the perplexity of     that span should increase when the entity mention is omitted.

We see that the median perplexity of a span increases by 32.2\% when the entity is removed, indicating that these spans are indeed related to the entity. Moreover, removing the distance-based criterion for span selection decreases the perplexity change to 25.9\%. These results indicate that our selected spans are correlated with the entity. This gap test was performed only for analysis and we do not use any model-based data filtering.

\renewcommand{\arraystretch}{1}
\begin{table}[t]
	\centering
	\footnotesize
	\setlength{\tabcolsep}{4pt}
	\begin{tabular}{l  r  r r r  }
		\toprule
		\multicolumn{1}{l }{ } & \# Sent. & \# Ent. & $m_q$ Span Len. & |Span V.| \\
  		\midrule
	$\text{LAMA}_{\text{TREx}}$   & 34k & 29,488 & 1.0 & 2,017  \\
        ECBD        & 35k & 2,106  & 2.9 & 19,542 \\
        \textsc{Popular} & 8k & 1,910 & 2.9 &  8,644\\
	
		\bottomrule 
	\end{tabular}
	\caption{Data statistics. |Span V.| means the vocabulary size of masked spans. Initial release of the data sample equal number of masked sentences per year (2017-21). }
	\label{tab:data-stats}
	\vspace{-11pt}
\end{table}

\vspace{-4pt}
\paragraph{Dataset Statistics}
Table~\ref{tab:per-type} shows the statistics and examples of ODIE, split by entity types. While our entity set does not comprehensively capture all entities originating in that year, it contains a diverse set of entities, ranging from events, products to organizations. One notably missing entity category is people; it is hard to pin down an origination year because of the significant gap between birth year and the year someone became prominent. 

Table~\ref{tab:data-stats} reports statistics on our cloze task data and existing probe dataset~\cite{petroni-etal-2019-language}. While containing fewer entities, our dataset exhibits much richer vocabulary (19K vs.~2K), demonstrating diverse knowledge it covers. We split this data into dev and test sets by entities (i.e., no shared entities between dev and test). To balance out the data sizes across the groups, we sample 4k examples for each year group, yielding 35k examples in total (approx. 20k for dev and 20k for test). Earlier dates contain a larger set of entities (599 entities for 2017 compared to 158 entities for 2021) as entities are continuously updated in Wikidata. In other words, many entities originated in 2021 have not been yet added to Wikidata. 
We sample the same number of NP spans and random spans. Within the NP spans, 35\% of them are proper noun phrases. 




\begin{figure*}
    \centering
    \includegraphics[width=1.0\linewidth]{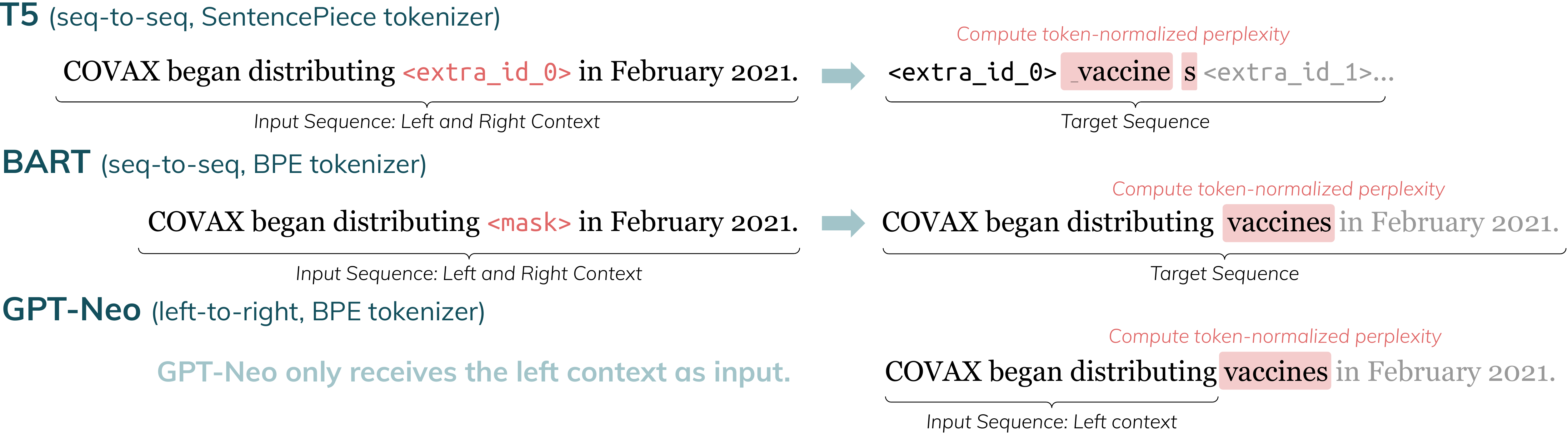}
    \caption{Perplexity computation over the masked span with three different modeling paradigms.}
    \label{fig:span}
    \vspace{-10pt}
\end{figure*}


\section{Experiments}

\paragraph{Setup}

We evaluate T5-large~\cite{T5}, BART-large~\cite{lewis-etal-2020-bart}, and GPT-Neo~\cite{gpt-neo} on our dataset in the zero-shot setting where the model parameters are fixed. In addition to the original masked sentence (\textsc{Original}), we feed three modified masked sentences. \textsc{No Ent} replaces the entity mention span with a generic string ``the entity.'' 
\textsc{Random Def.} prepends a definition sentence of a randomly selected entity. \textsc{Definition} prepends the first sentence of the entity's Wikipedia article to the cloze sentence.

We evaluate these models on the subsets split by year as well as a set of popular entities. Note that the entities in the 2020 and 2021 subsets are unseen for T5 and BART. Most entities in the 2020 and 2021 subsets are unseen to GPT-Neo, but its training data (the Pile \cite{gao2020pile}) does include the March 2020 English Wikipedia dump. In our experiments, we group the 2020 and 2021 subsets together as they consist of ``unseen'' entities. Similarly, we group the 2017, 2018, and 2019 subsets whose entities are ``seen'' during pre-training. See Appendix~\ref{sec:appendix_perplexity_year} for perplexity per year.

\renewcommand{\arraystretch}{1}
\begin{table}[h]
	\centering
    \footnotesize
	\setlength{\tabcolsep}{4.4pt}
	\begin{tabular}{l c c c }
		\toprule
		\multicolumn{1}{c}{} & \multicolumn{1}{c}{\textsc{Popular}} & \multicolumn{1}{c}{{\scriptsize \textsc{2017-2019}}} &\multicolumn{1}{c}{{ \scriptsize \textsc{2020-2021}}}\\

		\midrule
		\multicolumn{1}{l}{\scriptsize \textcolor{darkgray}{Type: seq-to-seq}} & \multicolumn{1}{c}{\textbf{\small T5 Large}} & \multicolumn{2}{r}{\scriptsize \textcolor{darkgray}{Size: 770M}} \\
		\midrule
		 \textsc{Original}               & 13.02 & 15.39 & 19.43 \\
		 \textsc{No Ent}                 & 18.28 & 22.35 & 26.69 \\
		 \textsc{Random Def.}            & 12.10 & 14.33 & 17.34 \\
		 \textsc{Definition}             & 11.04 & 11.73 & 13.60 \\
		 \arrayrulecolor{lightgray}\midrule\arrayrulecolor{black}
		 $\Delta$ ({\scriptsize \textsc{Orig.} $\rightarrow$ \textsc{Rand.}}) & \:-0.92 & \:-1.06 & \:-2.09 \\
		 $\Delta$ ({\scriptsize \textsc{Orig.} $\rightarrow$ \textsc{Def.}}) & \:-1.98 & \:-3.66 & \:-5.83 \\
		\midrule
		\multicolumn{1}{l}{\scriptsize \textcolor{darkgray}{Type: seq-to-seq}} & \multicolumn{1}{c}{\textbf{\small BART Large}}  & \multicolumn{2}{r}{\scriptsize \textcolor{darkgray}{Size: 406M}}\\
		\midrule
		 \textsc{Original}               & 22.70 & 21.09 & 28.79 \\
		 \textsc{No Ent}                 & 33.33 & 30.56 & 39.25 \\
		 \textsc{Random Def.}            & 27.69 & 25.59 & 33.74 \\
		 \textsc{Definition}             & 21.10 & 17.66 & 22.00 \\
		 \arrayrulecolor{lightgray}\midrule\arrayrulecolor{black}
		 $\Delta$ ({\scriptsize \textsc{Orig.} $\rightarrow$ \textsc{Rand.}}) & +4.99 & +4.50 & +4.95 \\
		 $\Delta$ ({\scriptsize \textsc{Orig.} $\rightarrow$ \textsc{Def.}}) & \:-1.60 & \:-3.43 & \:-6.79 \\
		\midrule
		\multicolumn{1}{l}{\scriptsize \textcolor{darkgray}{Type: left-to-right}} &\multicolumn{1}{c}{\textbf{\small GPT-Neo}}  & \multicolumn{2}{r}{\scriptsize \textcolor{darkgray}{Size: 1.3B}} \\
		\midrule
		 \textsc{Original}               & 28.61 & 27.81 & 33.36 \\
		 \textsc{No Ent}                 & 54.01 & 51.46 & 54.81 \\
		 \textsc{Random Def.}            & 39.46 & 41.03 & 45.92 \\
		 \textsc{Definition}             & 23.19 & 19.09 & 22.33 \\
		 \arrayrulecolor{lightgray}\midrule\arrayrulecolor{black}
		 $\Delta$ ({\scriptsize \textsc{Orig.} $\rightarrow$ \textsc{Rand.}}) & +10.85\:\: & +13.22\:\: & +12.56\:\: \\
		 $\Delta$ ({\scriptsize \textsc{Orig.} $\rightarrow$ \textsc{Def.}}) & \:-5.42 & \:-8.72 & -11.03\: \\
		 \bottomrule 
	\end{tabular}
	\caption{Results of T5, BART, and GPT-Neo on the test set, showing perplexity ($\downarrow$).}
	\label{tab:main-results-perplexity}
	\vspace{-12pt}
\end{table}
\vspace{-4pt}
\paragraph{Evaluation Metric} We compute token-normalized perplexity over the span as a proxy for entity knowledge stored in LMs. Each subset has different distribution of entity types (e.g., 2020 contains many COVID related entities and a lot less sports events compared to other years), and some frequent entities might contribute to perplexity excessively. To mitigate biases from particular entities, we first average negative log-likelihood (token normalized) over entities then average over examples. We follow the target sequence format used in LMs' pre-training tasks (see Figure~\ref{fig:span}).

Figure~\ref{fig:span} shows the perplexity computation. For left-to-right language models like GPT-Neo, we compute the perplexity of the span given the left context \emph{only}. T5 and BART, as seq2seq models, are able to also condition on the right context in their input; this makes perplexity values between these model classes not directly comparable (in addition to differences in tokenization and pre-training tasks). For T5 and BART, we condition on the input with a single mask. At decoding, for BART we initialize the decoder with the left context of the span and compute perplexity on the true span filler following this left context. For T5, we compute perplexity on the output span between the special tokens \texttt{<extra\_id\_0>} and \texttt{<extra\_id\_1>}.


\vspace{-4pt}
\paragraph{Results}

Table~\ref{tab:main-results-perplexity} reports perplexity (lower is better) on the test set that is split into three subsets: \textsc{Popular}, \textsc{2017-2019}, and \textsc{2020-2021}.
Note that absolute perplexity across years is sensitive to factors such as distribution of sentences or entity types; we thus focus on relative performance.

In all subsets, we observe two consistent trends across three LMs. (1) \textsc{No Ent} always degrades performance compared to \textsc{Original}. This result confirms that our masked spans are sensitive to the content of the entity span, although it is not conclusive proof of entity knowledge being required, as changing to ``the entity'' modifies other latent stylistic attributes that the LMs may be sensitive to. (2) \textsc{Definition} always boosts performance over \textsc{Original}, indicating that providing more information about entities helps to retrieve information distributed over LMs' parameters. \textsc{Random Def.} distracts BART and GPT-Neo but slightly improves T5 performance even though the additional information is taken from a random entity. This could be due to the model using different positional encodings as a result of having a definition, or LMs may select information if it is useful in some cases, leading the small gains.

\vspace{-4pt}
\paragraph{Performance on unseen entities} Recall that we consider \textsc{2020-2021} as unseen entities, and \textsc{2017-2019} and \textsc{Popular} as seen entities. All three LMs give higher perplexity on unseen entities, showing that the spans in \textsc{2020-2021} are relatively unexpected to the LMs. 

We further investigate the performance delta between \textsc{Original} and \textsc{Definition} per subset. For all three LMs, we see that the performace delta is relatively larger on \textsc{2020-2021}, indicating definition sentences are more useful on unseen entities.

Also, the performance delta on the popular entity set is notably smaller than \textsc{2020-2021} (compare T5 numbers: 13.02 $\rightarrow$ 11.04 for \textsc{Popular} versus 19.43 $\rightarrow$ 13.60 for \textsc{2020-2021}). This implies that LMs contain some prior knowledge about common entities they have observed before, and can use additional information about new entities or less frequent entities. How to inject knowledge requires further investigation.

\section{Use Cases}

We envision this dataset as being useful for general knowledge probing, as the real-world knowledge covered by the existing benchmarks is gradually outdated. With our framework, we can easily \textbf{update} datasets using the most recent knowledge sources with a controlled manner. Since the entity knowledge in our dataset is time-indexed, this is suitable for evaluating knowledge editing approaches \citep{Anton_Sinitsin_20, Chen_Zhu_2020, de-cao-etal-2021-editing, Eric_Mitchell_21, meng2022locating} and also continual knowledge learning approaches \citep{Joel_Jang_21}. Crucially, existing work studies whether these approaches can inject single facts, but not whether they can enable models to robustly support a broad range of new inferences about entities, like our dataset allows.

\section{Related Work}

Temporal mismatch/misalignment between large pre-trained LMs and real-world knowledge is an emerging research direction. \citet{Angeliki_Lazaridou_21} show that the corpus-level perplexity on documents from beyond LMs' training period becomes increasingly poor over time. \citet{Bhuwan_Dhingra_21} propose \textsc{TemporalLAMA}, which is based on time-dependent knowledge base triples (i.e., valid subject, relation, and object combinations given time). \textsc{SituatedQA} \cite{zhang-choi-2021-situatedqa} includes time-dependent QA examples. While these datasets primarily test temporal information about entities in the pre-training data, ECBD focuses on new entities which did not exist during pre-training. TemporalWiki \cite{Joel_Jang_2022} annotates new facts/entities based on the differences between Wikidata/English Wikipedia dumps, but does not necessarily reflect real-world changes during the time period (e.g., an ancient queen can be added to Wikidata in 2022). ECBD selects entities based on their origination date to align them with the real-world timeline.

Another line of work has looked at diachronic embeddings:
\cite{Derry_Tanti_Wijaya_2011,Yoon_Kim_2014,William_Hamilton_2016,Robert_Bamler_2017}, which can model changing meanings of words over time. Our setting focuses on introducing \emph{new} concepts rather than rewriting existing ones, but data similar to ECBD could be collected for new usages of existing words.

Although our dataset follows the widely-used cloze format, our focus is orthogonal to datasets like the Children’s Book Test \citep{Felix_Hill_16} and LAMBADA \citep{Paperno2016TheLD}, which come from fiction and do not cover real-world entities.  

\section{Conclusion}
In this paper, we present a dataset to understand language models' broad inferences about entities across time. We collect 43k cloze-style sentences associated with a time-indexed set of entities. We also perform analysis on our data set and show that handling completely unseen entities remains challenging for the current LMs.  

\section*{Acknowledgments}
This work was partially supported by NSF Grant IIS-1814522 and by the Air Force Research Laboratory (AFRL), Google Research Award, DARPA for the KAIROS program under agreement number FA8750-19-2-1003. The views and conclusions contained herein are those of the authors and should not be interpreted as necessarily representing the official policies, either expressed or implied, of DARPA, or the U.S. Government. The U.S. Government is authorized to reproduce and distribute reprints for governmental purposes notwithstanding any copyright annotation therein.


\bibliography{anthology,custom}
\bibliographystyle{acl_natbib}

\newpage
\appendix

\section{Examples of ECBD Sentences}
\label{sec:appendix_examples}

See Table~\ref{tab:examples-all} for examples of masked sentences in the ECBD data.

\renewcommand{\arraystretch}{1}
\begin{table*}[h]
	\centering
	\footnotesize
	\setlength{\tabcolsep}{4pt}
    \centering
	\begin{tabular}{l p{1.5cm}p{1.5cm}}
		\toprule
		\multicolumn{1}{c}{Masked Sentence} & \multicolumn{1}{c}{Span Type} & \multicolumn{1}{c}{Origin Year}\\
		\midrule
		 At 18:00 UTC on August 16, after {\bf Grace} exited the Dominican Republic, [MASK] were lifted.   & \multirow{2}{*}{\texttt{NP}} & \multirow{2}{*}{2021} \\
		 Answer: ``all tropical storm watches'' &  \\
		\midrule
		 {\bf AirTags} can be [MASK] the Find My app.   & \multirow{2}{*}{\texttt{RANDOM}} & \multirow{2}{*}{2021} \\
		 Answer: ``interacted with using'' &  \\
		\midrule
		 British tabloid ``The Sun'' is credited with the first headline use of `{\bf Megxit}' on [MASK] 2020.   & \multirow{2}{*}{\texttt{NP}} & \multirow{2}{*}{2020} \\
		 Answer: ``9 January'' &  \\
		\midrule
		 The {\bf iPhone SE} features an [MASK] a glass front and back. & \multirow{2}{*}{\texttt{RANDOM}} & \multirow{2}{*}{2020} \\
		 Answer: ``aluminum frame, paired with'' &  \\
		\midrule
		 The {\bf GPT-2} model has [MASK], and was trained on a dataset of 8 million web pages.   & \multirow{2}{*}{\texttt{NP}} & \multirow{2}{*}{2019} \\
		 Answer: ``1.5 billion parameters''  &  \\
		\midrule
		 The epicenter of the {\bf 2019 Albania earthquake} [MASK] kilometers from Tirana to the Northwest.   & \multirow{2}{*}{\texttt{RANDOM}} & \multirow{2}{*}{2019} \\
		  Answer: ``was about 30'' &  \\
		\midrule
		 On November 12, 2019, {\bf Maverick City Music} released [MASK], "Maverick City, Vol. 2".   & \multirow{2}{*}{\texttt{NP}} & \multirow{2}{*}{2018} \\
		 Answer: ``their follow-up EP'' &  \\
		\midrule
		 {\bf Austin FC} are the operators of a newly-[MASK].   & \multirow{2}{*}{\texttt{RANDOM}} & \multirow{2}{*}{2018} \\
		 Answer: ``built stadium at McKalla Place'' &  \\
		\midrule
		 The first quarter of {\bf Super Bowl LI} was [MASK] with each team punting twice. & \multirow{2}{*}{\texttt{NP}} & \multirow{2}{*}{2017} \\
		 Answer: ``a scoreless defensive match'' &  \\
		\midrule
		 {\bf Hurricane Irma} was the top Google searched term in [MASK] in 2017.   & \multirow{2}{*}{\texttt{RANDOM}} & \multirow{2}{*}{2017} \\
		 Answer: ``the U.S. and globally'' &  \\
		\bottomrule 
	\end{tabular}
\caption{Examples selected from the 2017-2021 subsets of ECBD.}
\label{tab:examples-all}
\end{table*}

 \section{Perplexity per year}
\label{sec:appendix_perplexity_year}

See Table~\ref{tab:main-results-perplexity-full} for a more fine-grained view of the results in Table~\ref{tab:main-results-perplexity}.

\renewcommand{\arraystretch}{1}
\begin{table*}[h]
	\centering
    \scriptsize
    \small
	\begin{tabular}{l c c c c c c}
		\toprule
		\multicolumn{1}{c}{} & \multicolumn{1}{c}{\textsc{Popular}} & \multicolumn{1}{c}{2017} & \multicolumn{1}{c}{2018} & \multicolumn{1}{c}{2019}&\multicolumn{1}{c}{2020}&\multicolumn{1}{c}{2021}\\

		\midrule
		\multicolumn{2}{l}{\scriptsize \textcolor{darkgray}{Type: seq-to-seq}} & \multicolumn{2}{c}{\textbf{\small T5 Large}} & \multicolumn{3}{r}{\scriptsize \textcolor{darkgray}{Size: 770M}} \\
		\midrule
		 \textsc{Original}               & 13.02 & 15.28 & 14.78 & 16.43 & 19.81 & 18.60  \\
		 \textsc{No Ent}                 & 18.28 & 22.28 & 21.70 & 23.35 & 28.41 & 23.26  \\
		 \textsc{Random Def.}            & 12.10 & 14.56 & 13.54 & 15.10 & 17.42 & 17.17  \\
		 \textsc{Definition}             & 11.04 & 12.27 & 10.76 & 12.34 & 14.07 & 12.61  \\
		 \arrayrulecolor{lightgray}\midrule\arrayrulecolor{black}
		 $\Delta$(\textsc{Orig.} $\rightarrow$ \textsc{Rand.}) & \:-0.92 & \:-0.72 & \:-1.24 & \:-1.33 & \:-2.39 & \:-1.43	\\
		 $\Delta$(\textsc{Orig.} $\rightarrow$ \textsc{Def.}) & \:-1.98 & \:-3.01 & \:-4.02 & \:-4.09 & \:-5.74 & \:-5.99 \\
		\midrule
		\multicolumn{2}{l}{\scriptsize \textcolor{darkgray}{Type: seq-to-seq}} & \multicolumn{2}{c}{\textbf{\small BART Large}}  & \multicolumn{3}{r}{\scriptsize \textcolor{darkgray}{Size: 406M}}\\
		\midrule
		 \textsc{Original}               & 22.70 & 22.74 & 19.52 & 21.00 & 28.03 & 30.53  \\
		 \textsc{No Ent}                 & 33.33 & 33.58 & 28.25 & 29.67 & 39.56 & 38.57  \\
		 \textsc{Random Def.}            & 27.69 & 27.11 & 23.80 & 25.96 & 32.41 & 36.86  \\
		 \textsc{Definition}             & 21.01 & 18.97 & 16.58 & 17.35 & 22.12 & 21.72  \\
		 \arrayrulecolor{lightgray}\midrule\arrayrulecolor{black}
		 $\Delta$(\textsc{Orig.} $\rightarrow$ \textsc{Rand.}) & +4.99 & +4.37 & +4.28 & +4.96 & +4.38 & +6.33  \\
		 $\Delta$(\textsc{Orig.} $\rightarrow$ \textsc{Def.}) & \:-1.69 & \:-3.77 & \:-2.94 & \:-3.65 & \:-5.91 & \:-8.81  \\
		\midrule
		\multicolumn{2}{l}{\scriptsize \textcolor{darkgray}{Type: left-to-right}} &\multicolumn{2}{c}{\textbf{\small GPT-Neo}}  & \multicolumn{3}{r}{\scriptsize \textcolor{darkgray}{Size: 1.3B}} \\
		\midrule
		 \textsc{Original}               & 28.61 & 28.91 & 27.55 & 26.63 & 33.15 & 33.81  \\
		 \textsc{No Ent}                 & 54.01 & 52.88 & 53.95 & 46.44 & 53.89 & 57.61  \\
		 \textsc{Random Def.}            & 39.46 & 41.75 & 43.15 & 37.41 & 45.30 & 47.32  \\
		 \textsc{Definition}             & 23.19 & 20.47 & 18.00 & 18.68 & 22.17 & 22.69  \\
		 \arrayrulecolor{lightgray}\midrule\arrayrulecolor{black}
		 $\Delta$(\textsc{Orig.} $\rightarrow$ \textsc{Rand.}) & +10.85\:\: & +12.84\:\: & +15.6\:\: & +10.78\:\: & +12.15\:\: & +13.51\:\:  \\
		 $\Delta$(\textsc{Orig.} $\rightarrow$ \textsc{Def.}) & \:-5.42 & \:-8.44 & \:-9.55 & \:-7.95 & -10.98 & -11.12  \\
		 \bottomrule 
	\end{tabular}
	\caption{Results of T5, BART, and GPT-Neo on the test set, showing perplexity ($\downarrow$) for each subset.}
	\label{tab:main-results-perplexity-full}
	\vspace{-0pt}
\end{table*}

\section{Perplexity per span type}
\label{sec:appendix_perp}
See Table~\ref{tab:perplexity-dev} for a breakdown of the perplexity that T5 achieves on different types of spans, showing that random spans are generally higher perplexity than NP spans but that adding definitions can help both.

\renewcommand{\arraystretch}{1}
\begin{table*}[h]
	\centering
    \small
	\setlength{\tabcolsep}{4pt}
	\begin{tabular}{l c c c c c c c c c c}
		\toprule
		\multicolumn{1}{c}{} & \multicolumn{2}{c}{2017}& \multicolumn{2}{c}{2018} & \multicolumn{2}{c}{2019}&\multicolumn{2}{c}{2020}&\multicolumn{2}{c}{2021}\\
		\cmidrule(r){2-3} \cmidrule(r){4-5} \cmidrule(r){6-7} \cmidrule(r){8-9} \cmidrule(r){10-11}
		\multicolumn{1}{c}{Input Type}  & \multicolumn{1}{c}{\sc  NP} & \multicolumn{1}{c}{ \sc Rand} &  \multicolumn{1}{c}{\sc  NP} & \multicolumn{1}{c}{ \sc Rand} &  \multicolumn{1}{c}{\sc  NP} & \multicolumn{1}{c}{ \sc Rand} &  \multicolumn{1}{c}{\sc  NP} & \multicolumn{1}{c}{ \sc Rand} &  \multicolumn{1}{c}{\sc  NP} & \multicolumn{1}{c}{ \sc Rand}\\
		\midrule
		 \textsc{Original}       & 5.86 & 7.33 & 5.81 & 7.51 & 6.11 & 7.29 & 5.92 & 7.63 & 6.23 & 7.31 \\
		 \textsc{No Ent}    & 5.90 & 8.02 & 5.78 & 8.56 & 5.99 & 8.31 & 6.75 & 9.36 & 7.28 & 9.21 \\
		 \textsc{Random Def.} & 5.59 & 6.60 & 5.54 & 6.84 & 5.77 & 6.60 & 5.70 & 6.98 & 6.01 & 6.65 \\
		 \textsc{Definition}        & 4.96 & 5.98 & 4.98 & 6.02 & 5.12 & 5.85 & 5.14 & 6.13 & 5.13 & 5.82 \\
		 \arrayrulecolor{lightgray}\midrule\arrayrulecolor{black}
        $\Delta$(\textsc{Orig.} $\rightarrow$ \textsc{Rand.}) & -0.27 & -0.73 & -0.27 & -0.67 & -0.34 & -0.69 & -0.22 & -0.65 & -0.22 & -0.66 \\
        $\Delta$(\textsc{Orig.} $\rightarrow$ \textsc{Def.}) & -0.90 & -1.35 & -0.83 & -1.49 & -0.99 & -1.44 & -0.78 & -1.50 & -1.10 & -1.49 \\
		\bottomrule 
	\end{tabular}
	\caption{Results of T5 model (pre-trained with data from 2019) on the dev set with perplexity ($\downarrow$) per span type.}
	\label{tab:perplexity-dev}
	\vspace{-0pt}
\end{table*}

\section{Recall@10}
\label{sec:appendix_recall}

LMs can be evaluated on \textbf{recall@10}, i.e., a binary score indicating if model's top ten predictions contains the gold masked span $m_y$. For T5, we first generate sequences using beam search (we choose beam size = 100 in our experiments). Then we take the top ten unique sequences and extract the text spans between \texttt{<extra\_id\_0>} and \texttt{<extra\_id\_1>} as predictions.  Table~\ref{tab:recall-dev-year} reports recall@10 on each subset. Table~\ref{tab:recall-dev-span} list recall@10 per span type for each subset.

We only explore recall on T5, since it is not obvious how to compute it for the other two models.
For BART, we can extract the predicted span by aligning the model's prediction with the gold context, assuming that it starts to copy from the input right context at some point. However, in some cases, we found that the generated right context does not match with the gold right context; it's unclear how to be handle this. For GPT-Neo, since it is a left-to-right LM, extracting the predicted span would require conditioning on the span length, which is information that T5 does not have access to. As a result, we do not report recall@10 for these models.

\renewcommand{\arraystretch}{1}
\begin{table*}[h]
	\centering
    \small
	\begin{tabular}{l c c c c c c}
		\toprule
		\multicolumn{1}{c}{} & \multicolumn{1}{c}{\textsc{Popular}} & \multicolumn{1}{c}{2017} & \multicolumn{1}{c}{2018} & \multicolumn{1}{c}{2019}&\multicolumn{1}{c}{2020}&\multicolumn{1}{c}{2021}\\
		\midrule
		\multicolumn{2}{l}{\scriptsize \textcolor{darkgray}{Type: seq-to-seq}} & \multicolumn{2}{c}{\textbf{\small T5 Large}} & \multicolumn{3}{r}{\scriptsize \textcolor{darkgray}{Size: 770M}} \\
		\midrule
		 \textsc{Original}               & 28.2 & 25.4 & 27.4 & 27.7 & 20.8 & 23.0 \\
		 \textsc{No Ent}                 & 23.8 & 21.6 & 23.2 & 23.7 & 19.5 & 21.5  \\
		 \textsc{Random Def.}            & 28.4 & 24.3 & 28.5 & 26.8 & 21.4 & 23.2 \\
		 \textsc{Definition}             & 29.3 & 28.4 & 31.8 & 28.2 & 24.8 & 26.1 \\
		 \arrayrulecolor{lightgray}\midrule\arrayrulecolor{black}
		 $\Delta$(\textsc{Orig.} $\rightarrow$ \textsc{Rand.}) & +0.2 & -1.1 & +1.1 & \:-0.9 & +0.6 & +0.2\\
		 $\Delta$(\textsc{Orig.} $\rightarrow$ \textsc{Def.}) & +1.1 & +3.0 & +4.4 & +0.5 & +4.0 & +3.1 \\
		 \bottomrule 
	\end{tabular}
	\caption{Results of T5 on the test set, showing recall@10 ($\uparrow$) for each subset.}
	\label{tab:recall-dev-year}
	\vspace{-0pt}
\end{table*}

\renewcommand{\arraystretch}{1}
\begin{table*}[h]
	\centering
	\small
	\setlength{\tabcolsep}{4pt}
	\begin{tabular}{l c c c c c c c c c c }
		\toprule
		\multicolumn{1}{c}{} & \multicolumn{2}{c}{2017}& \multicolumn{2}{c}{2018} & \multicolumn{2}{c}{2019}&\multicolumn{2}{c}{2020}&\multicolumn{2}{c}{2021}\\
		\cmidrule(r){2-3} \cmidrule(r){4-5} \cmidrule(r){6-7} \cmidrule(r){8-9} \cmidrule(r){10-11}
		\multicolumn{1}{c}{Input Type}  & \multicolumn{1}{c}{\sc  NP} & \multicolumn{1}{c}{ \sc Rand} &  \multicolumn{1}{c}{\sc  NP} & \multicolumn{1}{c}{ \sc Rand} &  \multicolumn{1}{c}{\sc  NP} & \multicolumn{1}{c}{ \sc Rand} &  \multicolumn{1}{c}{\sc  NP} & \multicolumn{1}{c}{ \sc Rand} &  \multicolumn{1}{c}{\sc  NP} & \multicolumn{1}{c}{ \sc Rand}\\
		\midrule
		\multicolumn{2}{l}{\scriptsize \textcolor{darkgray}{Type: seq-to-seq}} & \multicolumn{8}{c}{\textbf{\small T5 Large}} & \multicolumn{1}{r}{\scriptsize \textcolor{darkgray}{Size: 770M}} \\
		\midrule
		 \textsc{Original}       & 30.3 & 20.0 & 31.8 & 20.2 & 29.3 & 22.0 & 30.1 & 19.8 & 29.3 & 21.6 \\
		 \textsc{No Ent}    & 27.2 & 18.8 & 28.1 & 16.7 & 26.2 & 18.1 & 26.8 & 16.7 & 25.9 & 18.2 \\
		 \textsc{Random Def.} & 31.8 & 20.8 & 32.8 & 19.9 & 29.8 & 21.6 & 31.3 & 20.5 & 29.5 & 21.6 \\
		 \textsc{Definition}        & 34.1 & 22.8 & 35.9 & 22.8 & 33.0 & 24.9 & 33.7 & 23.0 & 32.7 & 25.2 \\
		 \arrayrulecolor{lightgray}\midrule\arrayrulecolor{black}
		 $\Delta$(\textsc{Orig.} $\rightarrow$ \textsc{Rand.}) & +1.5 & +0.8 & +1.0 & \:-0.3 & +0.5 & \:-0.4 & +1.2 & +0.7 & +0.2 & +0.0 \\
         $\Delta$(\textsc{Orig.} $\rightarrow$ \textsc{Def.} & +3.8 & +2.8 & +4.1 & +2.6 & +3.7 & +2.9 & +3.6 & +3.2 & +3.4 & +3.6 \\ 
		\bottomrule 
	\end{tabular}
	\caption{Results of T5 model (pre-trained with data from 2019) on the dev set with recall@10 ($\uparrow$) per span type.}
	\label{tab:recall-dev-span}
	\vspace{-0pt}
\end{table*}

\section{Data Licensing}

The Wikipedia text we used is licensed under CC BY-SA. Our use of Wikipedia, constructing a dataset which we will make publicly available under the same license, is consistent with the terms of the license.

\section{Computational Resources}

All experiments were conducted using an NVIDIA Quadro RTX 8000. We only evaluate existing models on our datasets and did not do any finetuning. One evaluation experiment typically takes 15  minutes to complete. For T5 experiments, we use Hugging Face's Transformer package \citep{wolf-etal-2020-transformers}.

\end{document}